\documentclass[5pt]{article}

\usepackage[letterpaper]{geometry}
\usepackage{spconf,amsmath,epsfig}
\usepackage{enumitem}

\usepackage{fancyhdr}
\begin{document}\sloppy
% Example definitions.
% --------------------
\def\x{{\mathbf x}}
\def\L{{\cal L}}

% Title.
% ------
\title{SALIENCY DEEP EMBEDDING FOR AURORA IMAGE SEARCH}
%
% Single address.
% ---------------
\name{Xi~Yang$^1$,~Xinbo~Gao$^2$, Bin Song$^1$, Nannan Wang$^1$ and Dong Yang$^3$\thanks{This work was supported in part by the National Natural Science Foundation of China (61602355, 61432014 and U1605252), the China Post-Doctoral Science Foundation (2016M590926), the Shaanxi Province Natural Science Foundation (2017JQ6007 and 2017JM6085) and Shaanxi Province Post-Doctoral Science Foundation.}}
\address{\footnotesize
$^1$State Key Laboratory of Integrated Services Networks, School of Telecommunications Engineering, Xidian University, China \\
\footnotesize
$^2$State Key Laboratory of Integrated Services Networks, School of Electronic Engineering, Xidian University, China\\
\footnotesize
$^3$Xi'an Institute of Space Radio Technology, China}

\maketitle

\begin{abstract}
Deep neural networks have achieved remarkable success in the field of image search. However, the state-of-the-art algorithms are trained and tested for natural images captured with ordinary cameras. In this paper, we aim to explore a new search method for images captured with circular fisheye lens, especially the aurora images. To reduce the interference from uninformative regions and focus on the most interested regions, we propose a saliency proposal network (SPN) to replace the region proposal network (RPN) in the recent Mask R-CNN. In our SPN, the centers of the anchors are not distributed in a rectangular meshing manner, but exhibit spherical distortion. Additionally, the directions of the anchors are along the deformation lines perpendicular to the magnetic meridian, which perfectly accords with the imaging principle of circular fisheye lens. Extensive experiments are performed on the big aurora data, demonstrating the superiority of our method in both search accuracy and efficiency.
\end{abstract}
\begin{keywords}
Aurora image search, convolutional neural network, saliency proposal network
\end{keywords}
\section{Introduction}
The powerful learning capacity of deep neural networks, especially the convolutional neutral network (CNN), has promoted the development in various computer tasks \cite{CNN}. In the field of image search, researchers introduce the CNN as an effective feature extraction tool into the content-based image retrieval (CBIR) framework. Although the existing CNN-based methods greatly improve the search accuracy compared with the SIFT-based methods, the images they process are captured with ordinary cameras without imaging distortion.

In practice, there are still a lot of images captured with circular fisheye lens with spherical aberration. Especially in the natural science field, phenomena in the sky are imaged from observation stations via the all sky imagers (ASI) to achieve bigger imaging vision. Thus, applying the CNN-based methods for the search of images captured with circular fisheye lens is a new task and of great significance.

\begin{figure}
\begin{center}
\includegraphics[width=0.9\linewidth]{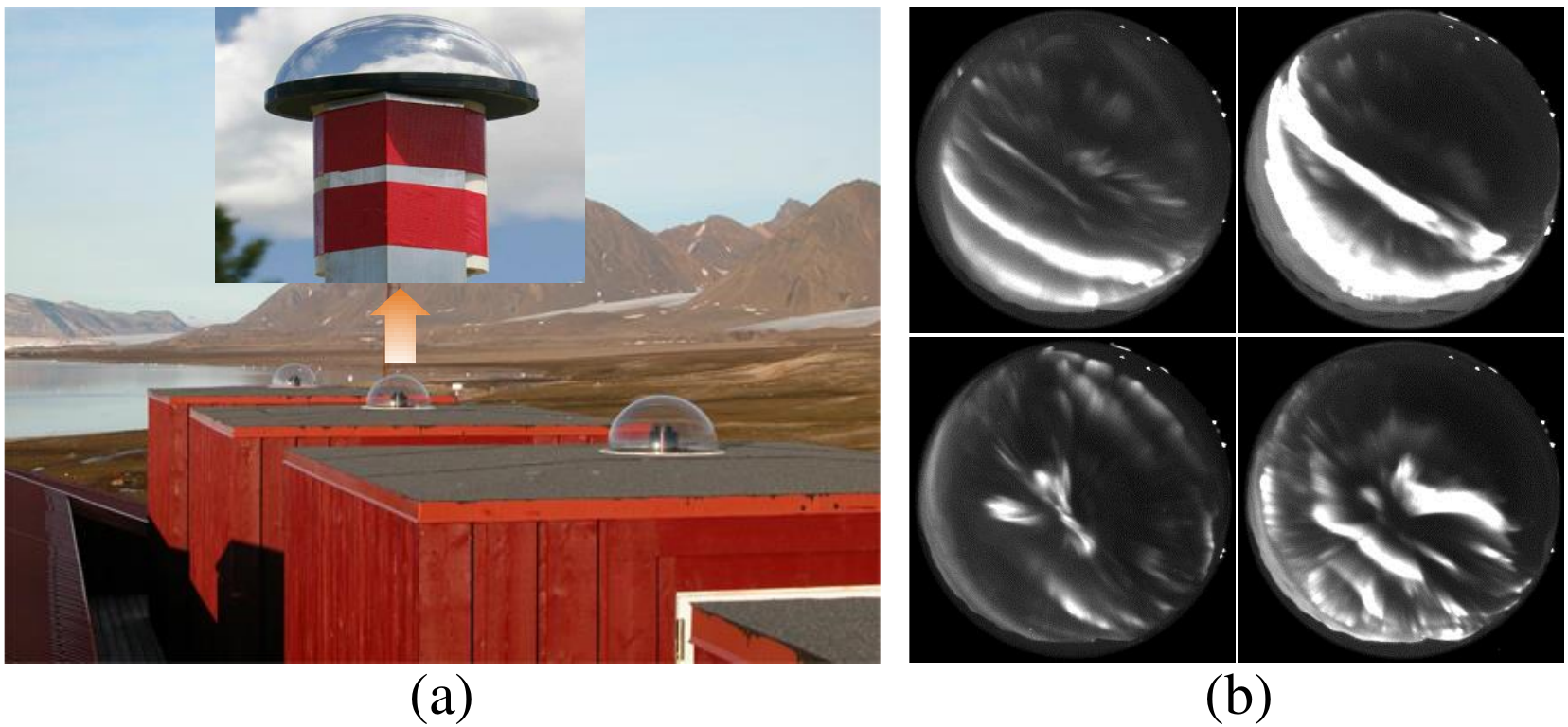}
\caption{\label{fig1}Related images of the ASI in YRS and the captured aurora images. (a) ASI in YRS. (b) Example aurora images.}
\end{center}
\vspace{-0.5cm}
\end{figure}

In this paper, we select the aurora images as a typical example to present our search method for circular fisheye lens. Generally, aurora is a natural light occurred in the high latitude regions caused by the collision of solar wind and particles in earth's magnetic field. As one of the most important natural phenomenon, scientists utilize multiple ways to capture it. In specific, the Yellow River Station (YRS) located at geographic coordinates 78.921N, 11.931E uses the ASIs to capture aurora occurred in the sky above, and the output image is in grayscale with size $512\times 512$. Fig. \ref{fig1} shows related images of the ASI in YRS and the captured aurora images.

Traditional aurora image search is conducted based on the visual observation. This manual way is time consuming and easily contaminated with visual fatigue. Hence, Yang \textit{et. al.} proposed a polar embedding (PE) model to realize automatic aurora image search. The PE model \cite{PE} leveraged the bag-of-words (BoW) framework \cite{BoW} and extracted the SIFT feature and deep local binary pattern (DLBP) feature for each keypoint. Although it achieved comparable performance in the big ASI aurora image dataset, its hand-crafted feature extraction and full domain computation limit the improvement of search accuracy and efficiency.

To this end, we propose a saliency deep embedding (SDE) model in this paper. On one hand, we exploit the powerful CNN model for feature extraction, and thus achieve a ``deep" understanding of aurora images. On the other hand, we present a ``saliency" domain computation way to reduce the memory cost and accelerate the online search time.

Specifically, the SDE model is inherited from the advanced Mask R-CNN \cite{Mask} and refined in consideration of the imaging principles of circular fisheye lens. Notably, Mask R-CNN is trained in the COCO dataset whose images are captured with ordinary cameras. Thus, the RPN is designed with rectangular meshing scheme, i.e., the centers of the sliding windows and anchors are distributed uniformly with horizontal surrounding boxes. In contrast, for ASI aurora images with spherical distortion, we revise the RPN as a SPN by combining the information of geomagnetic coordinates and circular imaging procedure. In our SPN, the locations of the anchors are distributed along the line of deformed latitude, while the shapes of anchors are determined based on the geometrical morphology of aurora structures. Also, the feature pyramid network (FPN) is applied in our SPN to achieve multi-scale feature representation. The proposed SDE model greatly facilitates the physics research, which liberates researchers from burdensome visual observation and is helpful for their further study of solar-terrestrial space.
\section{BACKGROUND}
\subsection{CNN-Based Image Search}
As one of the most famous deep neural network, CNN was proposed for the task of image classification with the last fully-connected layer exporting category labels. Commonly used models comprise AlexNet \cite{AlexNet}, VGG \cite{VGG}, ResNet \cite{ResNet} and its variants (e.g., ResNeXt \cite{ResNeXt}).

For the field of image search, CNN is utilized as a feature extraction tool while the fully-connected layers (e.g., FC6 in AlexNet) are first used as the high level semantic features. Also, region partition schemes, such as the spatial pyramid matching (SPM) \cite{SPM}, are explored to introduce spatial information and generate regional CNN feature. Related work includes the multi-scale orderless pooling (MOP) model \cite{MOP} which cascades regional CNN features as a global signature, and the probabilistic analysis (PA) model \cite{PA} which combines the CNN features in global image and multi-scale regions together. Subsequently, researchers observe that outputs of the convolutional layers are capable of representing mid-level features and thus achieve remarkable performance, e.g., maximum activation of convolutions (MAC) model \cite{MAC}.

In practice, the search target is not always the whole image but the objects therein. Hence, it is necessary to review the CNN models for object recognition. One of the most famous model is the regions with CNN features (R-CNN) \cite{rcnn}. However, this approach is memory consuming because the region determination is outside the deep network. To this end, numerous improved models are presented to design a region generation module which can be trained jointly with the main CNN structure. This series of work encompasses Fast R-CNN \cite{fast-rcnn}, Faster R-CNN \cite{faster-rcnn}, YOLO \cite{YOLO}, SSD \cite{SSD}, YOLO2 \cite{YOLO2}, etc. Especially, in the Faster R-CNN, the region generation module is designed as a fully convolutional neural network called region proposal network (RPN). Recently, the Mask R-CNN is proposed to realize pixel-level image segmentation, object detection and recognition simultaneously. It utilizes the feature pyramid network (FPN) for RPN to achieve multi-scale representation. Also, the replacement of RoIPool as RoIAlign promotes the precision of location projection between feature maps and the original image, thus achieving amazing performance.
\subsection{Characteristic of the ASI aurora image}
The ASI aurora images are captured from the ground to the sky above by circular fisheye lens with expanded field of view. Hence, the contents in ASI aurora image are distributed non-uniformity. Specifically, based on the imaging principle of circular fisheye lens, the peripheral zone exhibits more serious deformation than the central zone. Additionally, because of the gap between geomagnetic and geophysical coordinates in YRS, there is an offset angle $\varphi$ from the horizontal center line to the connection line of magnetic north (M.N.) and magnetic south (M.S.), i.e., magnetic meridian.

\begin{figure}
\begin{center}
\includegraphics[width=0.75\linewidth]{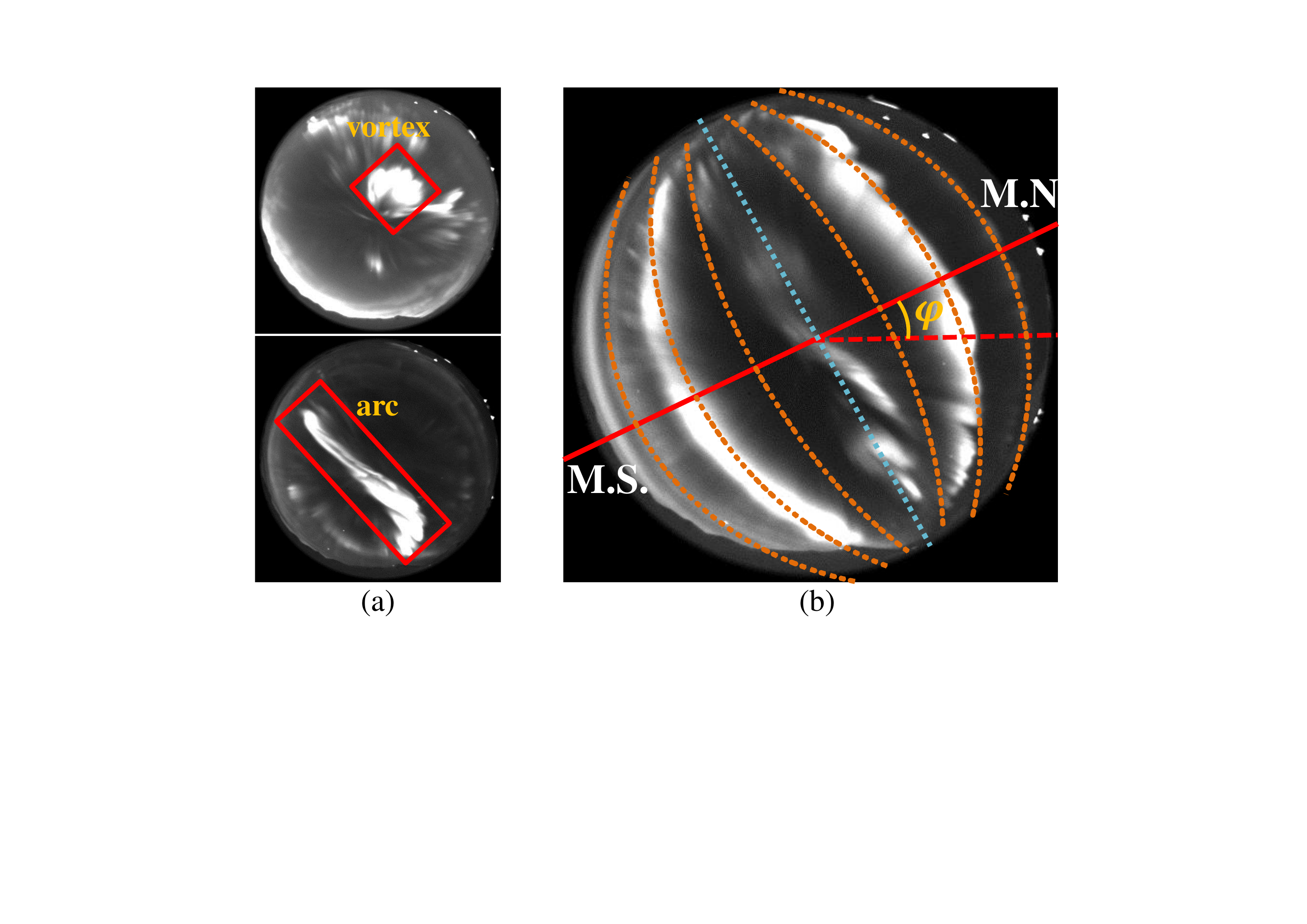}
\caption{\label{fig2}Characteristic of the ASI aurora image. (a) Auroral structures. (b) Deformation lines and magnetic meridian.}
\end{center}
\vspace{-0.5cm}
\end{figure}

Furthermore, the key factor affecting aurora image search results is the ``arc" or ``vortex" structures therein (as illustrated in Fig. \ref{fig2}(a)). If images in the dataset contain similar auroral structures with the query image, they can be exported as search results. Generally, as shown in Fig. \ref{fig2}(b) these auroral structures are located along the deformation lines (marked with orange dashed lines) perpendicular to the magnetic meridian (marked with red solid line).

\section{THE PROPOSED METHOD}
\begin{figure*}
\begin{center}
\includegraphics[width=0.65\linewidth]{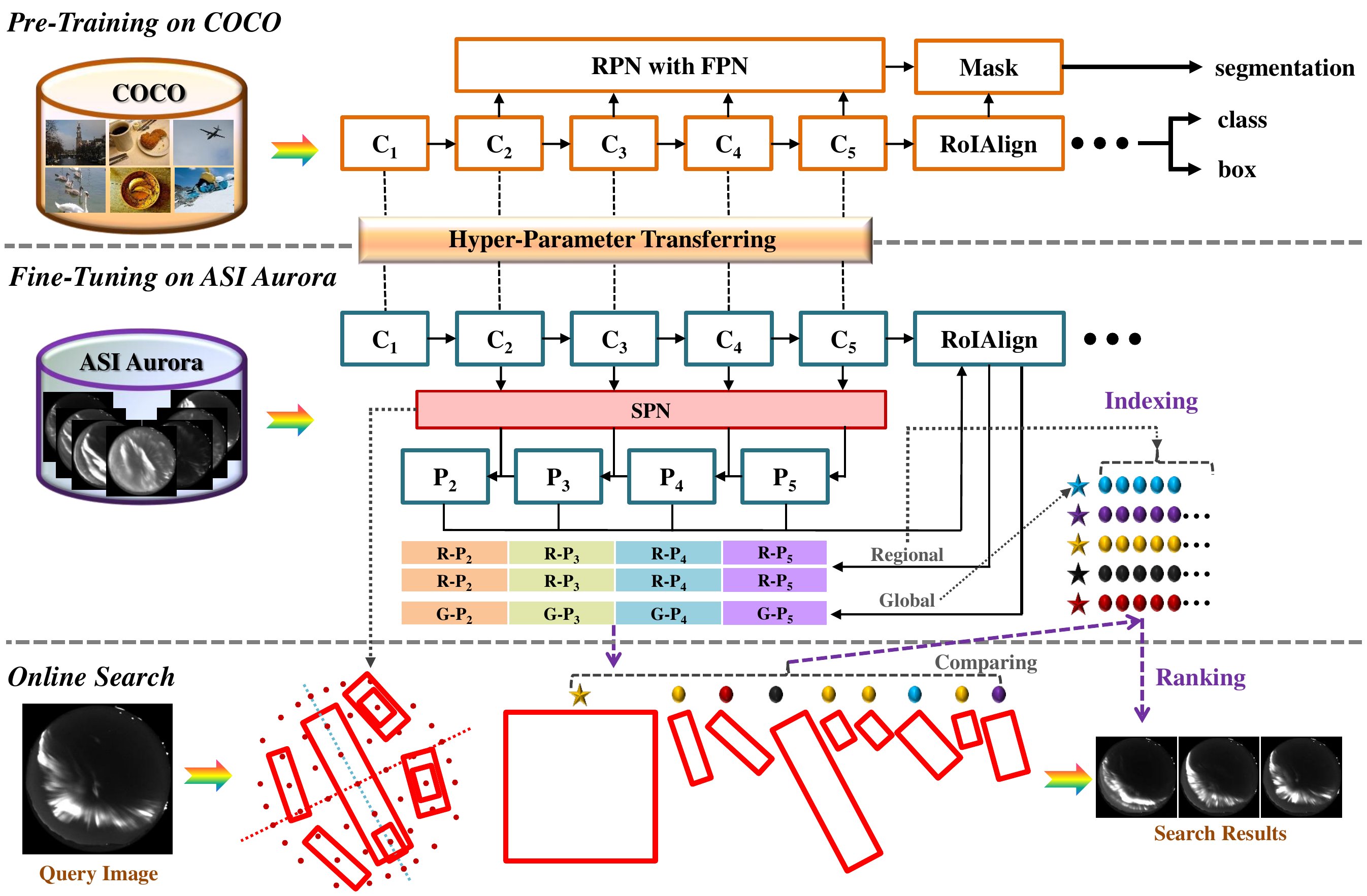}
\caption{\label{fig3}The diagram of the proposed SDE model.}
\end{center}
\vspace{-0.5cm}
\end{figure*}

This paper refines the Mask R-CNN model based on the characteristic of ASI aurora image, and thus proposes a SDE model for aurora image search. Our model focuses on saliency regions with informative auroral structures and presents a SPN to replace the traditional RPN. We first review the proposed method, and then explain the main innovations in detail.
\subsection{Overview of the proposed SDE model}
As illustrated in Fig. \ref{fig3}, the proposed SDE model is composed of three parts, i.e., pre-training on COCO, fine-tuning on ASI aurora and online search.

\textbf{\textit{Pre-Training on COCO}}. To inherit the strong power of existing CNN structure, we prefer the Mask R-CNN trained on COCO dataset as our pre-training model. Generally, the backbone architecture in Mask R-CNN is the Faster R-CNN with ResNet/ResNeXt and FPN. To preserve exact spatial locations in the feature maps, Mask R-CNN presents a RoIAlign layer which achieves prominent performance. In practice, we conduct hyper-parameters transferring on the convolutional layers (abbreviated as ${C_2}$ to ${C_5}$) as the initial setting for our SDE model, while the RPN module is discarded to redesign.

\textbf{\textit{Fine-Tuning on ASI Aurora}}. Since images in the COCO dataset are captured with ordinary cameras, we perform fine-tuning on our ASI aurora images to conform to the properties of circular fisheye lens. At first, data augmentation is conducted to make training data big enough. Then, hyper-parameters are fine-tuned for adapting our aurora dataset. Notably, the RPN is replaced as a SPN module. Here, the proposed SPN determines the locations of anchors based on the deformation lines and geomagnetic lines. While the directions and shapes of the bounding boxes around the anchors are trained individually. Subsequently, the FPN is leveraged to achieve multi-scale feature, and the RoIAlign layer is applied to realize precise location mapping between feature maps and the original image. Finally, top   regions are outputted with their features normalized to the same length. Together with the global CNN feature, all features related to each image are saved and the offline indexing is accomplished.

\textbf{\textit{Online Search}}. When given a query image, same operations are conducted to extract multi-scale CNN feature. Afterwards, referring to the indexing table, the global CNN features are first compared to compute a global similarity score. For each regional CNN feature in the query image, we scan all   regional CNN features of the compared image to get a regional similarity score. After processing all images in the dataset with the outputted similarity scores, search result is determined with the highest score regarded as the most similar one.
\subsection{SPN layer for feature extraction}
The proposed SDE model extracts both global and regional CNN features for searching, the regional CNN feature describes the specific auroral structure in an image, while the global CNN feature supplies the distribution of these auroral structures. In practice, only images achieve high similarity scores on both global feature and regional features can be treated as a similar result.

For the global CNN feature, we directly cascade the outputs of FPN $\{ {P_i}\}  = \{ {P_2},{P_3},{P_4},{P_5}\}$ without the region generation module. For the regional CNN feature, we propose a SPN layer to determine region proposals and directly link the SPN to FPN for generating multi-scale features. By considering the imaging principle of circular fisheye lens and the characteristic of the ASI aurora image, our SPN is mainly composed of three parts, i.e., circular anchors determination, region proposals detection and multi-scale feature extraction.

\begin{figure}
\begin{center}
\includegraphics[width=0.4\linewidth]{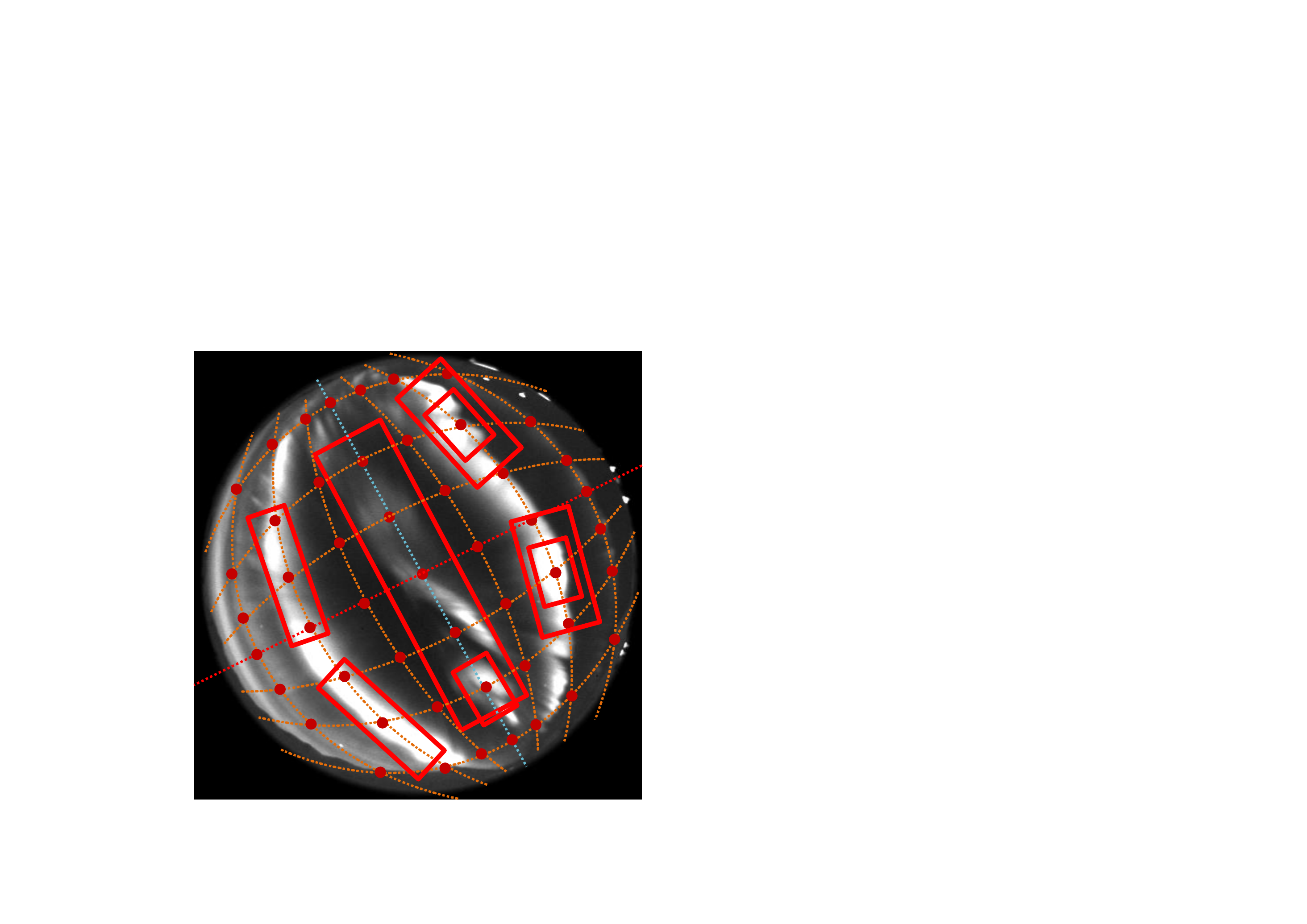}
\caption{\label{fig4}Circular anchors and region proposals determined by SPN. The circular anchors are marked with red dots and the corresponding region proposals are marked with red solid bounding boxes.}
\end{center}
\vspace{-0.5cm}
\end{figure}

\textbf{\textit{Circular Anchors Determination}}. Traditional R-CNN based methods regard the intersections of rectangular meshing as anchors, which is not suitable for images captured with circular fisheye lens. By analyzing the characteristic of ASI aurora image, our SPN considers both radial distortion and equatorial distortion. With the camera parameters and location information, the SPN draws the longitude and latitude deformation lines while treats their intersections as our circular anchors (see Fig. \ref{fig4}). If the magnetic meridian is divided into $l$ segments, all the deformation lines are cut into $l$ parts, thus resulting in ${l^2}$ anchors. It can be seen that our circular anchors not only accord with the shape of auroral structure, but reflect the information of geomagnetic latitude and longitude as well, which is capable of promoting the physical study in the future.

\textbf{\textit{Region Proposals Detection}}. Following the idea of Mask R-CNN, we generate bounding boxes centered at the circular anchors with different length-width ratios and scales. As illustrated in Fig.4, the directions of the bounding boxes are not horizontal as traditional methods, but along the deformation lines perpendicular to magnetic meridian. Hence, the resulting bounding boxes are able to enclose auroral structures with minimum area. Inspired by the work of YOLO2, we conduct $K$-means clustering to ASI aurora image dataset with regions labeled by polar experts. In practice, with the increase of clustering numbers $K$, the average IOU scores first increase rapidly and then attain a high value with small improvement. To strike a balance between precision and complexity, we finally choose $K=6$ bounding boxes priors. Thanks to these superior priors, the trained SPN is easy to learn and predict promising detections. Also, the FPN scheme is leveraged to achieve multi-scale region proposals generated from different pyramid layer ${P_i}$. Finally, region proposals with high probability carrying auroral structures are exported.

\textbf{\textit{Multi-Scale Feature Extraction}}. For each region proposal, we directly project its location to the corresponding feature maps with RoIAlign layer which is much more precise than the previous RoIPool. Afterwards, for each region proposal in each pyramid layer ${P_i}$, max-pooling is performed on the related tensors in feature maps and thus results in a feature vector $\{ R - {P_2},R - {P_3},R - {P_4},R - {P_5}\} $. By scanning all the region proposals in all pyramid layers, the regional CNN features in multi-scale are extracted. The same operations are conducted for the global image to obtain the global CNN feature $\{ G - {P_2},G - {P_3},G - {P_4},G - {P_5}\} $. Here, each $R - {P_i}$ or $G - {P_i}$ is set to 256-D. Consequently, the multi-scale feature we extracted covers multi-level sematic information.
\subsection{Indexing and querying}
For the ASI aurora dataset composing of $D$ images, the proposed SPN outputs ${r_d}$ regional CNN feature vectors and one global CNN feature vector for each image. Here, ${r_d}$ is depends on the number of auroral structures in the $d$th image. Hereafter, we save the image ID and the corresponding multi-scale CNN features into the indexing table.

In the stage of online search, we leverage the proposed SPN to extract CNN features for the query image following the same steps used in the offline training procedure. Also, the exported CNN feature in multi-scale consists of one global part and ${r_q}$ regional parts. By measuring the similarities in both global and regional parts between the query image $Q$ and image in the ASI aurora dataset $I$, the total similarity score can be computed as $SS(Q,I) = S{S^G}(Q,I) + S{S^R}(Q,I)$, where $S{S^G}(Q,I)$ and $S{S^R}(Q,I)$ are the global similarity score and regional similarity score, respectively, i.e.,
\begin{equation}\label{1}
S{S^G}(Q,I) = \frac{1}{{1 + dis({G^Q},{G^I})}},
\end{equation}
\begin{equation}\label{2}
dis({G^Q},{G^I}) = \sum\limits_{i = 2}^5 {d({G^Q} - {P_i},{G^I} - {P_i})},
\end{equation}
\begin{equation}\label{3}
S{S^R}(Q,I) = \frac{1}{{1 + \frac{1}{{{r_q}}}\sum\limits_{r = 1}^{{r_q}} {dis(R_r^Q,{R^I})} }},
\end{equation}
\begin{equation}\label{4}
dis(R_r^Q,{R^I}) = \frac{1}{{{r_d}}}\sum\limits_{r' = 1}^{{r_d}} {\sum\limits_{i = 2}^5 {d(R_r^Q - {P_i},R_{r'}^I - {P_i})} }.
\end{equation}
Here, $dis({G^Q},{G^I})$ is the distance between the global CNN feature in $Q$ and $I$, and it is actually the addition of the Euclidean distances between each $G - {P_i} (i = 2, \cdots 5)$. For the regional CNN features, $dis(R_r^Q,{R^I})$ is the average value of Euclidean distance addition between each $R - {P_i} (i = 2, \cdots 5)$ of one region in $Q$ and all other ${r_d}$ regions in $I$. By scanning all ${r_q}$ regions in the query image, its average value of the addition of $dis(R_r^Q,{R^I})$ can be computed. The lower of this value, the higher of the $S{S^R}(Q,I)$, indicating that $Q$ and $I$ are more similar in region proposals.

Notably, the regional similarity score $S{S^R}(Q,I)$ measures the matching degree of every auroral structure, while the global similarity score $S{S^G}(Q,I)$ represents the matching degree of the distributions of auroral structures. Only images similar on both global distribution and regional detailed structures can be treated as satisfying results. Finally, we rank the candidate images based on the total similarity score from highest to lowest.
\section{EXPERIMENTS ON AURORA IMAGE SEARCH}
To demonstrate the effectiveness of the proposed SDE model, extensive experiments are conducted on the ASI aurora image dataset. Our dataset are labeled with aurora experts, and they manually construct a query dataset ``ASI8K" including 10 special categories with 800 images for each one \cite{PE}. Other datasets with different sizes comprise ASI14K, ASI100K, ASI500K and ASI1M.
\subsection{Effectiveness of the SPN layer}
The major innovation of our SDE model is the SPN layer designed for images captured with circular fisheye lens. Our SPN layer determines the locations of circular anchors and the corresponding region proposals. To prove the effectiveness of these two aspects, we compare the rectangular meshing for anchors location determination and the horizontal direction for region proposals. Both of them are widely used in the RPN layer of CNN-based object detection methods.

Table \ref{table1} gives the comparison of mAPs using traditional RPN layer and the proposed SPN layer on datasets with increasing sizes. Noting that the datasets are abbreviated as their sizes (e.g., 8K for ASI8K), RA and CA stands for anchors with rectangular meshing and circular meshing, while HD and DD stands for anchors with horizontal direction and deformational direction.

It can be seen that the traditional RPN layer, which is actually the RA+HD approach, obtains the lowest mAP due to the inappropriate region proposals it determined. In contrast, CA+HD improves the mAP by 3.28\% on average, demonstrating the effectiveness of the circular anchors determination. Subsequently, CA+DD, which is actually the proposed SPN layer, further improves the mAP by 5.95\% on average and achieves the highest accuracy, indicating that the DD is superior than HD. Since the DD approach is based on the locations of CA, there is no RA+DD approach. Therefore, we can conclude that our SPN layer greatly promotes the performance of aurora image search.

\begin{table}
\small
\centering
\caption{\label{table1} Comparison of mAPs(\%) using traditional RPN layer and the proposed SPN layer.}
\begin{tabular}{cccccc}
\hline\hline
Methods		&8K	&14K	&100K	&500K	&1M	\\
\hline
RA+HD	&66.72	&65.12	&63.21	&61.01	&55.21\\
CA+HD	&70.14	&68.15	&65.47	&64.89	&59.01\\
CA+DD	&\textbf{73.16}	&\textbf{72.10}	&\textbf{71.90}	&\textbf{70.42}	&\textbf{69.85}\\
\hline\hline
\end{tabular}
\vspace{-0.5cm}
\end{table}
\subsection{Importance of the multi-scale feature}
Our SDE model leverages the multi-scale CNN feature for indexing and querying. This ``multi-scale" lies in two aspects. On one hand, we adopt the FPN to fuse features from different pyramid layers with different levels of semantic information and different resolutions. On the other hand, we extract not only regional CNN features to compare the similarity between auroral structures, but also global CNN feature to measure the overall distribution of related structures.

Table \ref{table2} illustrates the comparison of mAPs using different scales of CNN features. The first four rows present results with single pyramid regional CNN feature, from which we can see that the $R-{P_2}$ performs a little better than other layers due to its finest semantic level combining the information of $C_2$ to $C_5$. By merging all pyramid regional layers, the ${R-P_i}$ improves the search accuracy significantly, demonstrating the effectiveness of the usage of FPN in our framework. Similar results can be observed in the single pyramid global CNN feature and their combination ${G-P_i}$, and their mAPs are lower than the regional ones because of the rough description. Here, results of $G-P_3$ and $G-P_4$ are omitted for simplicity. Remarkably, our multi-scale CNN feature ${R-Pi}\&{G-Pi}$ gathers the advantages of regional and global information in different layers, and thus achieves the highest mAP.
\begin{table}
\small
\centering
\caption{\label{table2} Comparison of mAPs(\%) using different scales of CNN features.}
\begin{tabular}{cccccc}
\hline\hline
CNN feature		&8K	&14K	&100K	&500K	&1M	\\
\hline
$R-P_2$	&68.41	&67.45	&66.32	&64.95	&62.54\\
$R-P_3$	&68.21	&67.55	&66.00	&64.21	&61.95\\
$R-P_4$	&67.12	&67.23	&65.49	&64.15	&61.25\\
$R-P_5$	&67.03	&66.98	&65.12	&63.95	&61.02\\
${R-P_i}$	&70.54	&69.51	&68.39	&66.83	&65.21\\
\hline
$G-P_2$	&66.25	&65.98	&64.21	&63.22	&60.98\\
$G-P_5$	&65.84	&64.32	&63.54	&61.11	&59.87\\
${G-P_i}$	&68.74	&67.24	&66.52	&64.81	&63.14\\
\hline
${R-P_i}\&{G-P_i}$	&\textbf{73.16}	&\textbf{72.10}	&\textbf{71.90}	&\textbf{70.42}	&\textbf{69.85}\\
\hline\hline
\end{tabular}
\vspace{-0.5cm}
\end{table}
\subsection{Comparison with the state-of-the-art methods}
We compare the proposed SDE model with the state-of-the-art methods including BoW (baseline) \cite{BoW}, PE \cite{PE}, MOP \cite{MOP}, PA \cite{PA}, MAC \cite{MAC}, R-CNN \cite{rcnn} and Mask R-CNN \cite{Mask}. All experiments are conducted on the same environment for ASI aurora image search. The comparison results are shown in Table \ref{table3}, and the first five columns are mAPs for datasets with increasing sizes, while the last column indicates the average query time for one image. By analyzing the comparison results, we can draw the following conclusions.

\begin{table}
\footnotesize
\centering
\caption{\label{table3} Comparison with the state-of-the-art methods in accuracy measured by mAP(\%) and efficiency measured by average query time (s).}
\begin{tabular}{ccccccc}
\hline\hline
Methods		&8K	&14K	&100K	&500K	&1M	 &time\\
\hline
BoW (baseline)	&49.01	&47.65	&46.65	&42.32	&39.09	&1.98\\
PE	&62.88	&61.09	&60.35	&58.31	&57.99	&0.82\\
MOP	&67.12	&66.49	&65.12	&64.58	&63.22	&2.32\\
PA	&68.00	&67.15	&66.87	&65.41	&64.98	&1.76\\
MAC	&65.58	&64.02	&61.58	&59.12	&58.87	&1.22\\
R-CNN	&66.54	&65.42	&64.85	&63.52	&62.12	&1.25\\
Mask R-CNN	&69.01	&68.47	&67.95	&66.24	&64.12	&0.62\\
SDE (our)	&\textbf{73.16}	&\textbf{72.10}	&\textbf{71.90}	&\textbf{70.42}	&\textbf{69.85} &\textbf{0.59}\\
\hline\hline
\end{tabular}
\vspace{-0.5cm}
\end{table}

\textbf{\textit{Accuracy}}. 1) The PE outperforms the BoW because of the introduction of polar meshing scheme. 2) Compared with the BoW and PE, other methods leveraging CNN features achieve higher values of mAP, demonstrating the powerful representational capacity of CNN feature. 3) In the CNN-based methods, the Mask R-CNN yields outstanding performance because of the strong network design and RoIAlign layer for accurate location projection. 4) Remarkably, the proposed SDE model attains the highest mAP in all datasets with different sizes. In particular, we gets satisfying mAP of 69.85\% in ASI1M, which is far ahead of other comparison methods.

\textbf{\textit{Efficiency}}. Due to the introduction of region proposal network, Mask R-CNN and the proposed SDE achieve favorable efficiency compared with other methods. The best value of our method lies in two aspects. 1) We only process the ``saliency" regions with abundant auroral structures, thus reducing the computational complexity compared with the full domain approach. 2) The circular anchors are more effective for aurora images, which allow us to leverage less region proposals without reducing the search accuracy.
\section{CONCLUSION}
This paper proposes a SDE model for aurora image search. By analyzing the imaging principle of circular fisheye lens, the SPN is presented into the Mask R-CNN framework to generate region proposals suitable for aurora images. The locations and directions of the obtained region proposals are along the deformation lines perpendicular to magnetic meridian, which facilitates physicists for their research on solar-terrestrial space. In practice, by ignoring the uninformative regions and only excavating the ¡°saliency¡± regions encompassing auroral structures, the computational complexity is greatly reduced without losing the accuracy. Additionally, to supplement the local similarity measured by SPN, the global CNN feature is extracted to evaluate the global similarity. Thus, our method improves the search accuracy with high efficiency.

In the future, we will apply our method to other images captured by unordinary cameras. Also, how to refine the feature saving approach to achieve efficient indexing and querying is also an interesting problem.

\bibliographystyle{IEEEbib}
\bibliography{SDE}

\end{document}